\documentclass[sn-aps]{sn-jnl}

\DeclareMathOperator*{\Mean}{Mean}
\DeclareMathOperator*{\argmax}{arg\hspace{0.8pt}max}
\DeclareMathOperator*{\argmin}{arg\hspace{0.8pt}min}
\newcommand{\mmid}{\! \mid \!}
\renewcommand*{\v}[1]{\boldsymbol{#1}}

\newcommand{\abs}[1]{\lvert #1 \rvert}

\begin{document}

\title[Deep reinforcement learning for the olfactory search POMDP]{Deep reinforcement learning for the olfactory search POMDP: a quantitative benchmark}

\author*[1]{\fnm{Aurore} \sur{Loisy}}\email{aurore.loisy@irphe.univ-mrs.fr}

\author*[2]{\fnm{Robin A.} \sur{Heinonen}}\email{robin@physics.ucsd.edu}


\affil[1]{Aix Marseille Univ, CNRS, Centrale Marseille, IRPHE, Marseille, France}

\affil[2]{Dept. Physics and INFN, University of Rome ``Tor Vergata'', Via della Ricerca Scientifica 1, 00133 Rome, Italy}

\abstract{The olfactory search POMDP (partially observable Markov decision process) is a sequential decision-making problem designed to mimic the task faced by insects searching for a source of odor in turbulence, and its solutions have applications to sniffer robots. As exact solutions are out of reach, the challenge consists in finding the best possible approximate solutions while keeping the computational cost reasonable. We provide a quantitative benchmarking of a solver based on deep reinforcement learning against traditional POMDP approximate solvers. We show that deep reinforcement learning is a competitive alternative to standard methods, in particular to generate lightweight policies suitable for robots.}


\maketitle

\section{Introduction}

Partially observable Markov decision processes (POMDPs) provide an elegant mathematical framework to model decision-making in the face of uncertainty \cite{Astrom1965,Smallwood1973,Cassandra1994}. They generalize MDPs (Markov decision processes) to situations where the agent has only access to partial information about the state of the world, for example through sensors. In real life, partial observability is the rule rather than the exception, and an important application of POMDPs is robot navigation \cite{Cassandra1996,Thrun2006book}. 

The olfactory search POMDP is a navigation problem where the agent must find a source of odor in a turbulent flow using information provided by odor detection events \cite{Vergassola2007}. This task is faced by insects searching for food or mates using their sense of smell \cite{Murlis1992,Vickers2000,Carde2021}, but also by sniffer robots used to locate gas leaks, land mines and explosives \cite{Russell1999book}. Far from a toy problem, the olfactory search POMDP reproduces the key features of odor detection in turbulence: sparsity and stochasticity \cite{Celani2014}. It can be used to assess and compare possible search strategies \cite{Loisy2022a} before implementing them in real robots \cite{Lochmatter2010thesis,Moraud2010,Martinez2013}. It also provides a tool to interpret the behavior of olfactory animals \cite{Calhoun2014,Voges2014}.

The POMDP framework rigorously models uncertainty that arises from partial observability. It allows the agent to compute a probability distribution over possible states of the world (possible source locations) and to update it as new sensory information (odor detections) arrives. This probability distribution, called the belief, is a sufficient statistic of the entire agent's history and completely describes the current uncertainty about the true state of the world (the true source location). Solving the POMDP means computing the optimal action to take as a function of the belief. This solution optimally balances exploration (acting to gain more information about the source location) and exploitation (acting to get closer to the current estimate of the source location).

The price to pay for this careful quantification of uncertainty is computational complexity. Finding the optimal strategy for a POMDP requires to solve a nonlinear functional equation, called the Bellman optimality equation, on the space of beliefs. This problem is computationally intractable and one must rely on approximate solvers \cite{Kurniawati2022,Kochenderfer2022book}. Most popular solvers are ``point-based'' and compute the solution by performing value iteration over a subset of beliefs \cite{Pineau2006,Shani2013}. A different approach was recently proposed in \cite{Loisy2022a}, where deep reinforcement learning techniques were adapted to the POMDP framework and applied to the olfactory search problem. While this approach was novel, it has not been compared to well-established techniques, which are also able to obtain good approximate solutions for this problem \cite{Heinonen2022arxiv}. 

In this paper we benchmark deep reinforcement learning against standard point-based solvers on the olfactory search POMDP. 
The POMDP and its (formal) optimal solution are described in Section~\ref{sec:pomdp}. The reinforcement learning approach to POMDP is presented in Section~\ref{sec:drl_and_solvers} and contrasted with existing point-based solvers. The benchmark methodologies and results are detailed in Section~\ref{sec:benchmark}. Conclusions are drawn in Section~\ref{sec:conclusion}.

\section{The olfactory search POMDP}
\label{sec:pomdp}

\begin{figure*}
\centering
\includegraphics[width=0.99\textwidth]{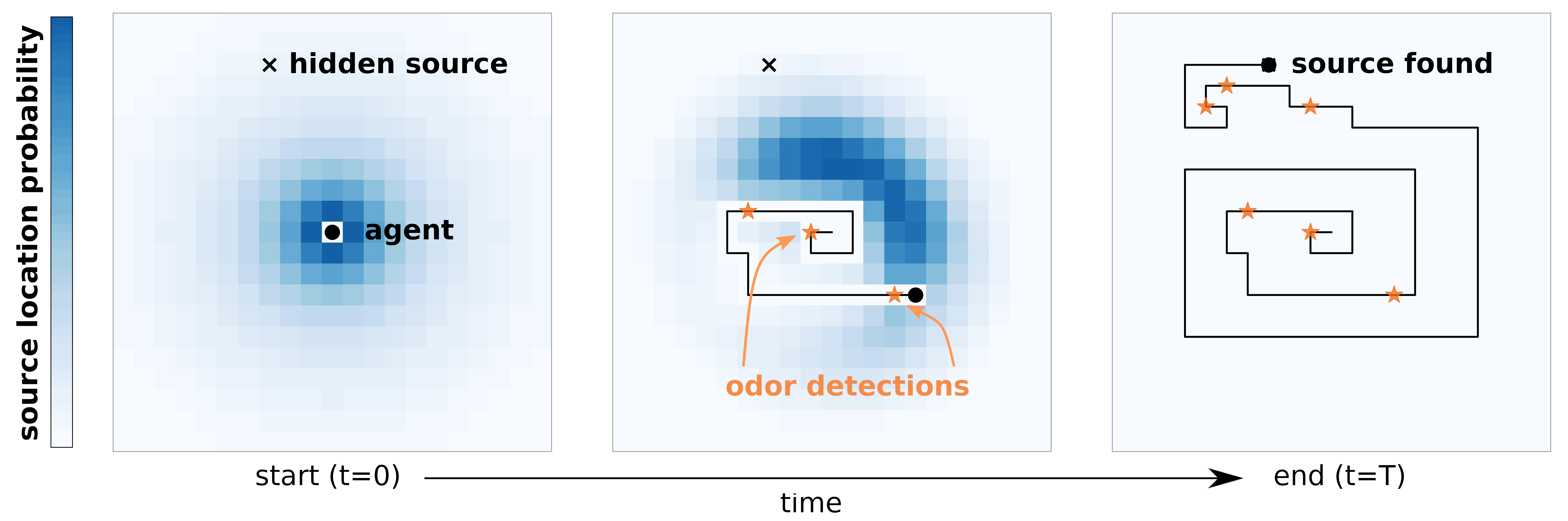}
\caption{Illustration of the olfactory search POMDP. In this POMDP, the agent must find a source of odor randomly hidden in a grid. At each step, the agent moves to a neighbor cell (action). It sniffs the air and has a small random chance of detecting an odor (observation). The closer it is to the source, the higher is the probability of detection. The search terminates when the agent enters the cell containing the source. The model used to generate odor detections (called ``hits'') is known, so the agent can maintain a belief (probability distribution over possible source locations) using Bayesian inference. We seek the policy (mapping from beliefs to actions) that minimizes the expected search duration.}
\label{fig:illustration_pomdp}
\end{figure*}

The olfactory search problem, illustrated in Fig.~\ref{fig:illustration_pomdp}, is a POMDP in which the agent must find a source of odor hidden in a 2D Cartesian grid. At each step, the agent moves to one of the four adjacent cells. If the source is located in this cell, the search is over. Otherwise, the agent receives a stochastic sensor measurement in the form of ``hits''. Hits represent odor particles detected by the agent. Their likelihood depends on the position of the agent with respect to the source. Therefore hits provide noisy information about the source location. The search continues until the agent enters the cell containing the source. We assume that the agent has a perfect memory and a perfect knowledge of the process that generates hits. The goal is to determine the strategy that the agent should follow in order to minimize the expected number of steps to find the source.

We now formally define this problem in the language of POMDPs. 
The state $s$ is defined as the relative position of the agent with respect to the source, and belongs to the set $\mathcal{S}$ built from all possible combinations of the source's and the agent's positions within a finite-size grid. The special state $s^\Omega = \v{0}$ is the terminal state where the agent is located in the cell containing the source.
The action $a$ is a move by the agent, and belongs to the set $\mathcal{A}=$\{`north,' `south,' `east,' `west'\}. As the agent executes an action $a$, it transitions deterministically to a new state $s$, receives a reward $r=-1$, and makes an observation $o$. Possible observations belong to the set $\mathcal{O} = \{\Omega, 0, 1, 2, \dots, h_{\text{max}} \}$.
If $s=s^\Omega$, the source is found: the agent receives the special observation $o=\Omega$ and the search terminates.
If $s\neq s^\Omega$, the agent receives an observation $o=h$ where $h \in \{ 0, 1, 2, \dots, h_{\text{max}} \}$ is a number of hits (the maximum number of hits $h_{\text{max}}$ will be specified later on). Hits represent odor detections and occur with conditional probability $\Pr(h \mmid s)$, which is constructed from a physical model of odor dispersion and detection in turbulence \cite{Vergassola2007} (cf. Appendix~\ref{app:obs_model}).

The agent does not have access to its current state. Instead, it maintains a probability distribution over $\mathcal{S}$, called belief and denoted $b(s)$, as an estimate of its state. At each step, after making an observation $o$, the belief is updated from $b$ to $b'$ using Bayes' rule
\begin{equation}
\label{eq:bayes}
 b'(s) = \dfrac{\Pr(o \mmid s) b(s)}{\displaystyle \sum_{s \in \mathcal{S}} \Pr(o \mmid s) b(s)}.
\end{equation}
This update can made more explicit depending on the nature of the observation.
If $o=\Omega$, the source is found and the update then yields $b'(s)=\delta(s^\Omega)$. We denote this special belief $b^\Omega$. Otherwise, $o=h$ and
\begin{equation}
 b'(s) = \dfrac{\Pr(h \mmid s) b(s)}{\displaystyle \sum_{s \in \mathcal{S}} \Pr(h \mmid s) b(s)} \qquad \forall s \neq s^\Omega
\end{equation}
with $b'(s^\Omega) = 0$, where $\Pr(h \mmid s)$ is given in Appendix~\ref{app:obs_model}.

The belief summarizes all the information brought by past observations and actions. It is a sufficient statistic over the agent's history: we can reason equally about beliefs as about agent histories without loss of information. The initial belief, $b_0$, is called the ``prior'' in the language of Bayesian probabilities and is somewhat arbitrary. Here, $b_0$ is drawn from a set of initial beliefs $\mathcal{B}_0$. This set is generated by assuming that the search starts with a detection ($h > 0$) in an infinite domain with a uniform source distribution. The motivation behind this initialization procedure is that the start of the search is not arbitrary, but is triggered at the instant when the agent is informed that there is source (as opposed to nothing) in the neighborhood. It also drastically reduces artificial effects due to the finite size of the search domain. The details of the initialization protocol are provided in Appendix~\ref{app:init} and \cite{Loisy2022a}.

The search proceeds as follows:
\begin{itemize}
    \item Initially
        \begin{itemize} 
         \item The initial belief $b_0$ is drawn randomly from $\mathcal{B}_0$.
         \item The state $s_0$ is drawn randomly from the distribution $b_0$.
        \end{itemize}
    \item At the $t^\text{th}$ step of the search
    \begin{enumerate}
    \item The agent chooses an action according to some policy $\pi$: $a_{t+1} = \pi(b_t)$.
    \item The agent moves deterministically to the neighbor cell associated with $a_{t+1}$. This move is associated with a negative unit reward: $r_{t+1}=-1$. The state is updated to $s_{t+1}$.
    \item The agent makes an observation $o_{t+1}$ and the belief is updated to $b_{t+1}$.
    \begin{itemize}
    \item If $b_{t+1} = b^\Omega$ (meaning that $o_{t+1} = \Omega$ and $s_{t+1} = s^\Omega$), the source is found and the search terminates.
    \item Otherwise, the search continues to step $t+1$.
    \end{itemize}
    \end{enumerate}
\end{itemize}
Each search (called an episode) is a sequence like this:
\begin{equation*}
b_0, a_1, r_1, s_1, o_1, b_1, a_2, r_2, s_2, o_2, b_2, \dots, b_{T-1}, a_T, r_T, s^\Omega, \Omega, b^\Omega.
\end{equation*}
and the cumulative reward of an episode is equal to minus the number of steps $T$ to termination: $\sum_{t=1}^T r_t = -T$.

The agent's behavior is controlled by the policy, denoted $\pi$, which maps each belief to an action: $a=\pi(b)$.
The performance of a policy $\pi$ is measured by $\mathbb{E}_{\pi} [T]$, the expected number of steps to reach the source. The expectation is taken over all possible sequences generated following policy $\pi$ starting from all possible initial states $s_0$. Solving the POMDP means finding the optimal policy $\pi^*$ that minimizes the expected duration of the search
\begin{equation}
 \pi^* = \argmin_\pi \mathbb{E}_{\pi} [T].
\end{equation}
The optimal policy can, at least formally, be determined from the solution of a recurrence equation known as the Bellman optimality equation as follows.

\begin{figure}
    \centering
	\includegraphics[height=3cm]{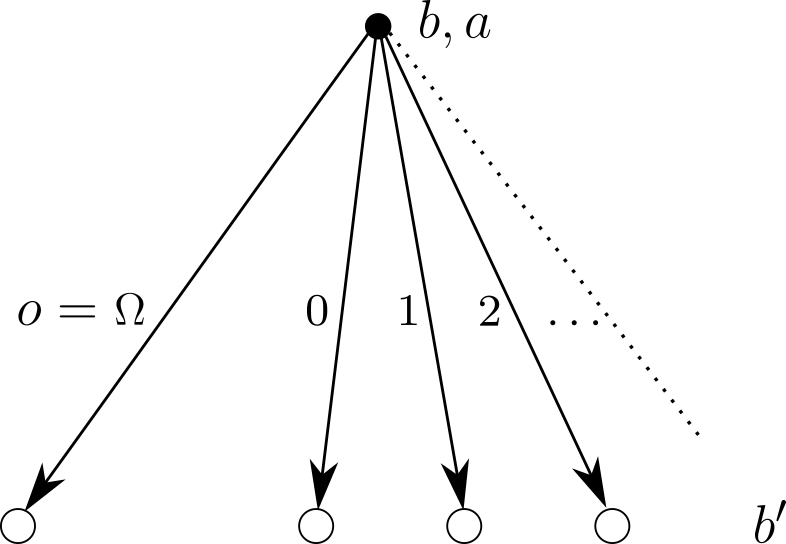}
    \caption{
    Tree of possible successor beliefs $b'$ starting from a belief $b$ and executing action $a$. Transitions from $b$ to $b'$ are determined by the observation $o$.
	\label{fig:successor_beliefs}
	}
\end{figure}

The optimal value function $v^*(b)$ of a belief $b$ is defined as the maximum, over all policies, of the expected cumulated reward when starting from this belief $b$. Here, the reward is a simple unit penalty at each step, so $v^*(b)$ is equal to the expected number of steps remaining to find the source up to a minus sign:
\begin{equation}
 v^*(b) = \max_\pi v^\pi(b) \qquad \text{where} \quad v^\pi(b) = - \mathbb{E}_{\pi} [T - t \mmid b_t = b].
\end{equation} 
The optimal value function satisfies the Bellman optimality equation:
\begin{equation}
\label{eq:optimal_Bellman}
    v^*(b) = -1 + \max_{a \in \mathcal{A}} \sum_{b' \in \mathcal{B}(b,a)} \Pr(b'\mmid b,a) v^*(b') \qquad  \forall b \neq b^\Omega
\end{equation}
with $v^*(b^\Omega) = 0$, where $\mathcal{B}(b,a)$ is the set of successor beliefs $b'$ reachable from a belief $b$ by executing action $a$ and where $\Pr(b'\mmid b,a)$ is the probability of transitioning from $b$ to $b'$ upon $a$. Possible transitions correspond to possible observations $o$, as illustrated in Fig.~\ref{fig:successor_beliefs}, and transition probabilities are given by $\Pr(o) = \sum_{s \in \mathcal{S}} \Pr(o \mmid s) b(s)$. Once a function solution to the Bellman optimality equation is found, the optimal policy consists in selecting the action that maximizes the expected optimal value:
\begin{equation}
 \pi^*(b) = \argmax_{a \in \mathcal{A}} \sum_{b' \in \mathcal{B}(b,a)} \Pr(b'\mmid b,a) v^*(b').
\end{equation}

The optimal value function cannot be computed exactly due to the size of the belief space. It can, however, be approximated, as explained in the next section.

\section{Approximate POMDP solvers}
\label{sec:drl_and_solvers}

Exactly solving the Bellman optimality equation of a POMDP is computationally intractable for any but the smallest problems, and a tremendous amount of effort has been devoted to the development of approximation methods \cite{Kurniawati2022,Kochenderfer2022book}. The general idea is to solve the Bellman optimality equation only for a set of sampled beliefs rather than for the entire belief space, thereby substantially reducing the complexity of the problem. In order to obtain good policies, it is key to sample a sufficiently representative set of beliefs; therefore, the sampling strategy is critical to the performance of the method.

In the following we present the two generic approaches that are compared in this paper: model-based deep reinforcement learning and point-based POMDP solvers.

\subsection{Model-based deep reinforcement learning}

\begin{figure*}
\centering
\includegraphics[width=0.99\textwidth]{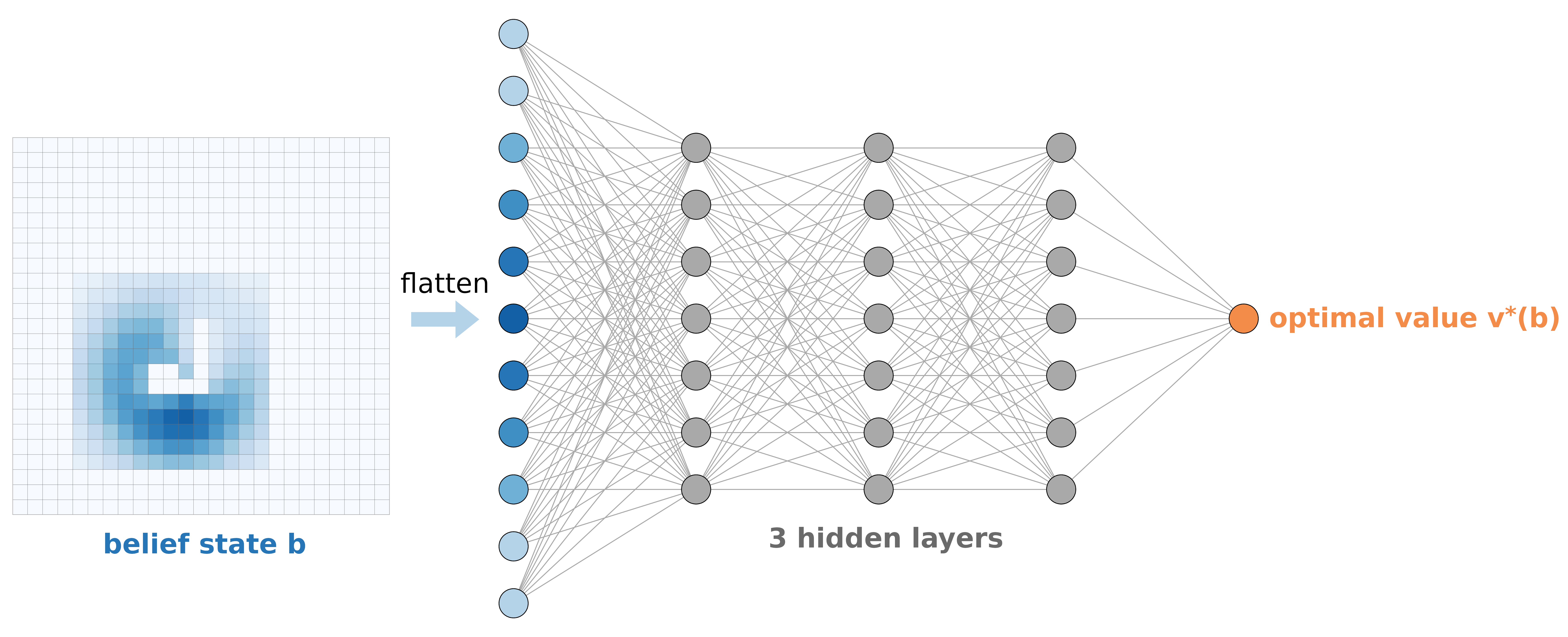}
\caption{Neural network approximation of the optimal value function used in model-based deep reinforcement learning. The weights are optimized so as to minimize the error on the Bellman optimality equation (Eq.~\ref{eq:approximate_optimal_Bellman}).}
\label{fig:nn_illustration}
\end{figure*}

The deep reinforcement learning approach to solving a POMDP \cite{Loisy2022a} consists in approximating the optimal value function by a deep neural network, and training the network to minimize the error on the Bellman optimality equation (Fig.~\ref{fig:nn_illustration}).

We denote by $\hat{v}(b;\v{w})$ the neural network approximation of $v^*(b)$ parameterized by weights $\v{w}$. The Bellman optimality equation for the approximate value function reads
\begin{equation}
\label{eq:approximate_optimal_Bellman}
    \hat{v}(b;\v{w}^*) = -1 + \max_{a \in \mathcal{A}} \sum_{b' \in \mathcal{B}(b,a)} \Pr(b'\mmid b,a) \hat{v}(b';\v{w}^*)  \qquad  \forall b \neq b^\Omega
\end{equation}
with $\hat{v}(b^\Omega;\v{w}^*) = 0$. The problem becomes that of computing the weights $\v{w}^*$ that minimize the residual error on Eq.~\ref{eq:approximate_optimal_Bellman}. This residual error, called the Bellman optimality error, reads
\begin{equation}
\label{eq:loss_definition}
 L(\v{w}) = \mathbb{E}_{b \sim \hat{\pi}} \left[ -1 + \max_{a \in \mathcal{A}} \sum_{b' \in \mathcal{B}(b,a)} \Pr(b'\mmid b,a) \hat{v}(b';\v{w}) - \hat{v}(b;\v{w}) \right]^2
\end{equation}
where the expectation is taken over beliefs $b$ visited when following the policy $\hat{\pi}$ derived from $\hat{v}$:
\begin{equation}
\label{eq:approximately_optimal_policy_definition}
    \hat{\pi}(b;\v{w}) = \argmax_{a \in \mathcal{A}} \sum_{b' \in \mathcal{B}(b,a)} \Pr(b'\mmid b,a) \hat{v}(b';\v{w}).
\end{equation}
Using neural network terminology, the functional $L(\v{w})$ is the ``loss function'' to minimize and ``training'' the network refers to the iterative update of the weights $\v{w}$ using stochastic gradient descent.

The intuition behind this deep reinforcement learning approach is the following.
At the beginning of the training, $\hat{v}$ is initialized with random weights $\v{w}$. As the consequence, the policy $\hat{\pi}$ used to collect beliefs, which is derived from $\hat{v}$, is random. At each training iteration, the weights are adjusted such that $\hat{v}$ becomes a better approximation of the true optimal value function on a collection of beliefs gathered by following $\hat{\pi}$ (Eq.~\ref{eq:loss_definition}). As $\hat{v}$ is improved, beliefs collected from $\hat{\pi}$ become more representative of the beliefs visited by the optimal policy. This allows to improve $\hat{v}$ even further. This iterative process continues until convergence to $\hat{v} \approx v^*$ and $\hat{\pi} \approx \pi^*$.

The training algorithm is a model-based version of DQN (Deep Q-Network) which relies on two stabilizing techniques to facilitate convergence (which in general is not guaranteed): experience replay and delayed target network \cite{Mnih2015}. It is model-based because it takes advantage of model knowledge: since the probability of transitioning from a belief $b$ to a successor belief $b'$ is known exactly, one can work directly with the value function rather than the action-value ``Q function'' in model-free reinforcement learning, and one can perform full backups (compute the sum over $b'$ in equations \ref{eq:approximate_optimal_Bellman}-\ref{eq:approximately_optimal_policy_definition}) rather than sample backups (estimates based on a single successor belief randomly sampled) in model-free reinforcement learning.
The complete algorithm is provided in Algorithm~\ref{alg:drl}.

\subsection{Standard point-based POMDP solvers}

The ``classical'' approach to approximating the optimal policy for POMDP is to use point-based value iteration (PBVI) \footnote{PBVI often refers to a specific algorithm introduced in Ref.~\cite{Pineau2003}, the first of its kind. For our purposes, it refers to the broader class of point-based algorithms for value iteration.} \cite{Pineau2003,Pineau2006,Shani2013}. PBVI approximates the optimal value function as piecewise-linear and convex, i.e.,\
\begin{equation}
\hat{v}(b;\Gamma) = \max_{\alpha\in\Gamma} \sum_{s \in \mathcal{S}} b(s) \cdot \alpha(s),
\end{equation}
for some collection $\Gamma$ of $\alpha$-vectors (it has been shown \cite{Sondik1971thesis} that $v^*$ can be arbitrarily well-approximated by such a function).
The challenge consists in constructing the set $\Gamma^*$ such that the Bellman optimality equation
\begin{equation}
\label{eq:Bellman_approx_discount}
    \hat{v}(b;\Gamma^*) = -1 + \gamma \max_{a \in \mathcal{A}} \sum_{b' \in \mathcal{B}(b,a)} \Pr(b'\mmid b,a) \hat{v}(b';\Gamma^*)  \qquad  \forall b \neq b^\Omega
\end{equation}
is solved ``at best'' for $\hat{v}$. Note that here we have introduced a discount factor $\gamma = 1 - \epsilon$ with $0 < \epsilon \ll 1$, which is required for PBVI. We refer to the Appendix~\ref{app:solver_pbvi} for more details.

It can be shown that when viewed as an operator acting on $\hat{v}$, the right-hand-side of Eq.~\ref{eq:Bellman_approx_discount} is a contraction and $\hat{v}(\cdot;\Gamma^*)$ is a fixed point of this operator. This property is the basis of value iteration, an iterative algorithm that proceeds as follows at the $n^{\text{th}}$ iteration:
\begin{equation}
\hat{v}(b,\Gamma^{n+1}) = - 1 + \gamma \max_{a \in \mathcal{A}} \sum_{b' \in \mathcal{B}(b,a)} \Pr(b' \mmid b,a) \hat{v}(b';\Gamma^n)
\end{equation}
for $b$ in a collection of beliefs. The set $\Gamma^{n+1}$ is built up by performing a ``backup'' operation on $b$: this generates a new $\alpha$-vector which improves the approximation of that belief's value (and presumably that of other beliefs close to it). We refer the reader to the related technical literature \cite{Shani2013} for more information on backups.

PBVI algorithms differ in how the beliefs to be backed up are chosen and the order in which they are backed up. Due to the size of the belief space, which is frequently very high-dimensional, choosing an efficient scheme is of critical importance.

Perseus \cite{Spaan2005} constructs its set of beliefs by collecting them along trajectories generated using a heuristic policy. Then the beliefs are backed up, one by one, until every belief satisfies $\hat{v}(b; \Gamma^{n+1}) \ge \hat{v}(b; \Gamma^n)$ (as a consequence of convexity, backups can only increase the estimated value of a belief). The order of the backups is either random or (our preference) in order of decreasing Bellman error \cite{Shani2008}. For very large POMDPs (as considered here) where the set of beliefs can only be a very small subset of the entire belief space, the quality of the heuristic used to sample beliefs is key: it must visit beliefs that are representative of the beliefs visited by the optimal policy.

Sarsop \cite{Kurniawati2008}, on the other hand, interleaves belief sampling and backups which allows it to be more parsimonious in its selection of beliefs. It tries to construct a tree of beliefs which are reachable from the initial belief $b_0$ by taking sequences of quasi-optimal actions, pruning branches corresponding to provably suboptimal actions. An advantage of this approach is that it maintains lower and upper bounds on the exact optimal value which progressively tighten as the algorithm proceeds. The algorithm stops when the distance between the bounds for $b_0$ is sufficiently small.

Perseus and Sarsop do not comprise an exhaustive list of PBVI algorithms for POMDPs. However, they are popular, and there is little interest in testing every available solver.

\section{Quantitative benchmark}
\label{sec:benchmark}

\subsection{Methods}
\label{sec:bench_methods}

\begin{table}
\begin{center}
\begin{tabular}{@{}lrrrrrrr@{}}
\toprule
case & grid size  & $\abs{\mathcal{S}}$ & $\abs{\mathcal{A}}$ & $h_{\text{max}}$ & $\abs{\mathcal{O}}$ & $\abs{\mathcal{B}_0}$ & $T_{\text{max}}$ \\
\midrule
isotropic, smaller domain & $19 \times 19$ & 1369       & 4 & 2  & 4 & 2 & 642 \\
isotropic, larger domain & $53 \times 53$ & 11025       & 4 & 3  & 5 & 3 & 2188 \\
windy, with detections & $81 \times 41$ & 13041         & 4 & 1  & 3 & 1 & 10000 \\
windy, almost no detections & $81 \times 41$ & 13041    & 4 & 1  & 3 & 1 & 10000 \\
\botrule
\end{tabular}
\begin{tabular}{@{}lrrr@{}}
\toprule
case & Mean(T) & P$_{99}$(T) & Mean(cum. hits) \\
\midrule
isotropic, smaller domain    & $\phantom{0}13.2$ & $79$ & $\phantom{0}1.7$ \\
isotropic, larger domain    & $\phantom{0}34\phantom{.0}$ & $152$ & $10\phantom{.0}$ \\
windy, with detections & $\phantom{0}63\phantom{.0}$ & $274$ & $\phantom{0}6\phantom{.0}$ \\
windy, almost no detections & $217\phantom{.0}$ & $1331$ & $\phantom{0}1.6$ \\
\botrule
\end{tabular}
\caption{Description of the four test cases used for benchmarking.
In the top table, we describe the size of the problem. The state is defined as the position of the agent relative to the source, so for a grid size $n_x \times n_y$ the number of possible states is $\abs{\mathcal{S}} = (2 n_x-1) (2 n_y - 1)$. The action space $\mathcal{A}$ consists of the 4 possible moves to adjacent cells. The observation space $\mathcal{O}$ is the union of the terminal observation $\Omega$ and of the set $\{0, 1, \dots, h_{\text{max}} \}$ of hit values that can be received at each step. $\mathcal{B}_0$ is the set of initial beliefs. $T_{\text{max}}$ is the maximum time allowed for the search (if the source is not found at $t=T_{\text{max}}$, the search is considered a 'failure').
In the bottom table, we provide information on search times and number of odor detections.
Mean(T) is the mean number of steps to find the source. P$_{99}$(T) is number of steps after which 99 \% of the sources are found. Mean(cum. hits) is the mean number of hits cumulated during the search. The reported values for Mean(T), P$_{99}$(T) and Mean(cum. hits) correspond to the best policy we obtained for each case.}
\label{tab:cases}%
\end{center}
\end{table}

\begin{figure*}
\centering
\includegraphics[width=0.99\textwidth]{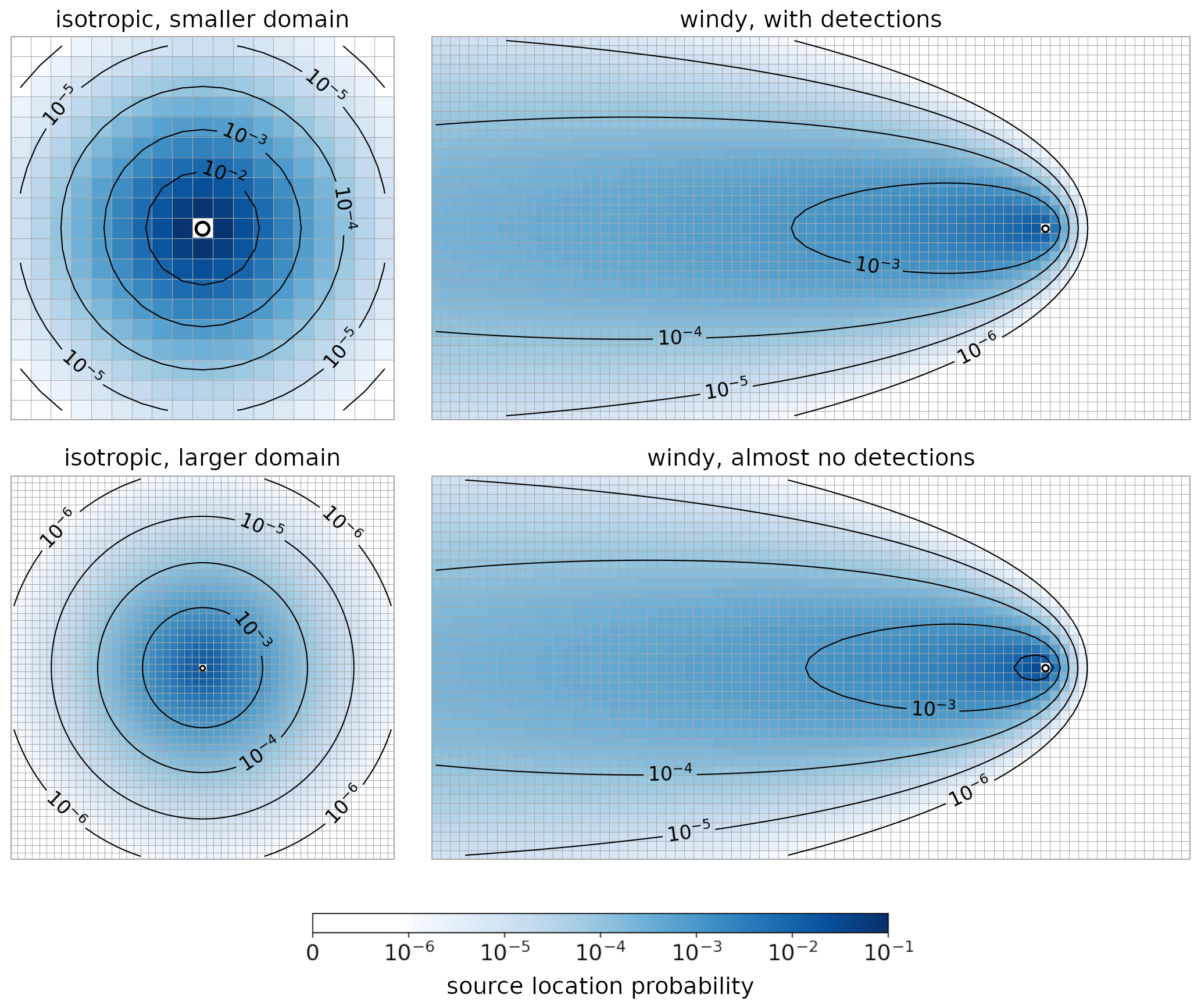}
\caption{Probability distribution of the source location for the four test cases. The agent's starting position is fixed, and is denoted by a circle. The source location is drawn randomly at the beginning of each search from this probability distribution. For the isotropic cases, the distribution shown here is actually the weighted sum of the different possible initial beliefs corresponding to the different possible initial hits. For the windy cases, a single initial belief is considered and corresponds to the probability distribution shown here.}
\label{fig:source_distribution_cases}
\end{figure*}

We consider four test cases which are described in Table~\ref{tab:cases}. In the first two test cases, the problem is isotropic: the search domain is a square grid and the agent starts the search at the center of the domain. Two different problem sizes are considered: a $19 \times 19$ grid (``smaller'') and a $53 \times 53$ grid (``larger''). In both cases the source emission rate is chosen such that occasional detections are likely to occur during the search. Possible hit values are integers between 0 and 2 or 3. The initial belief is drawn from a small set containing 2 or 3 initial beliefs corresponding with an initial nonzero hit. The setup is identical to that used in \cite{Loisy2022a}. In the last two test cases, the effect of a mean wind in the positive x-direction is accounted for: the domain is a rectangular grid ($81 \times 41$) and the agent starts downwind at position $(66,21)$. Two different source emission rates are considered, such that the search is performed either with occasional detections (``with detections'') or almost without any detections (``almost no detections''). Possible hit values are either 0 or 1. A single initial belief is used. This setup is similar to that used in \cite{Heinonen2022arxiv}. The probability distribution of possible source locations and the initial position of the agent are shown in Fig.~\ref{fig:source_distribution_cases} for each test case. Additional technical details on the setups are provided in Appendix~\ref{app:cases_details}. 

Approximately optimal policies have been computed using three different computational methods: deep reinforcement learning (DRL), Sarsop \cite{Kurniawati2008} and Perseus \cite{Spaan2005}. We briefly summarize our methodology in the remainder of this section; refer to Appendix~\ref{app:solvers} for further details.

The DRL method is very generic and has no theoretical restrictions on the type of POMDP it can solve, but involves a large number of hyperparameters. We found empirically that the quality of the solution shows very little sensitivity to most hyperparameters. Only the size of the neural network and the learning rate must be chosen in an appropriate manner (large enough and small enough, respectively). Based on extensive experiments by \cite{Loisy2022a}, we use a learning rate of 10$^{-3}$ and a fully connected neural network with 3 hidden layers of 512 neurons for the small isotropic case, and of 1024 neurons for all other cases. We found that increasing the network size further does not yield further improvements to the learned policy.

Sarsop and Perseus are two standard PBVI solvers \cite{Shani2013}. In principle, these solvers can only deal with discounted POMDPs (the olfactory search POMDP is undiscounted). In practice we found that good policies can be obtained for the undiscounted POMDP ($\gamma=1$) while solving for the discounted version of the problem ($\gamma<1$). An additional limitation of Sarsop is that it requires a single initial belief, while there are several ones in our isotropic test cases. As one initial belief is much more likely than other ones (cf. Appendix~\ref{app:init}), Sarsop was used considering only this initial belief. The out-of-the-box version of Perseus is unable to obtain good approximate solutions. Nevertheless, using a good heuristic (we use infotaxis \cite{Vergassola2007}) instead of a random policy to collect beliefs and using reward shaping allowed us to use Perseus for the olfactory search POMDP \cite{Heinonen2022arxiv}.

Policy evaluation has been performed with OTTO, a software dedicated to the olfactory search POMDP and designed for this purpose \cite{Loisy2022b}. This software has been augmented with the windy setup, which was not present in the original version, and adapted so that it can use policies computed with PBVI solvers. The augmented version of OTTO used for this paper can be found at \url{https://github.com/auroreloisy/otto-benchmark}.

As policies, in general, do not guarantee that the source will be always found, a search may never terminate. We prescribe a maximum search time $T_{\text{max}}$ for each case (cf. Table~\ref{tab:cases}), which is chosen much larger than the maximum time it takes a good policy to find the source. If the source is not found at $t=T_{\text{max}}$, the episode is marked as 'failure'. The performance of a policy is defined based on two metrics: $\Mean(\text{T})$, the mean time to find the source conditioned on the fact that the source is found, and $\Pr(\text{failure})$, the probability that a search ends by a failure.

\subsection{Results}
\label{sec:bench_results}

\begin{figure*}
\centering
\includegraphics[width=0.99\textwidth]{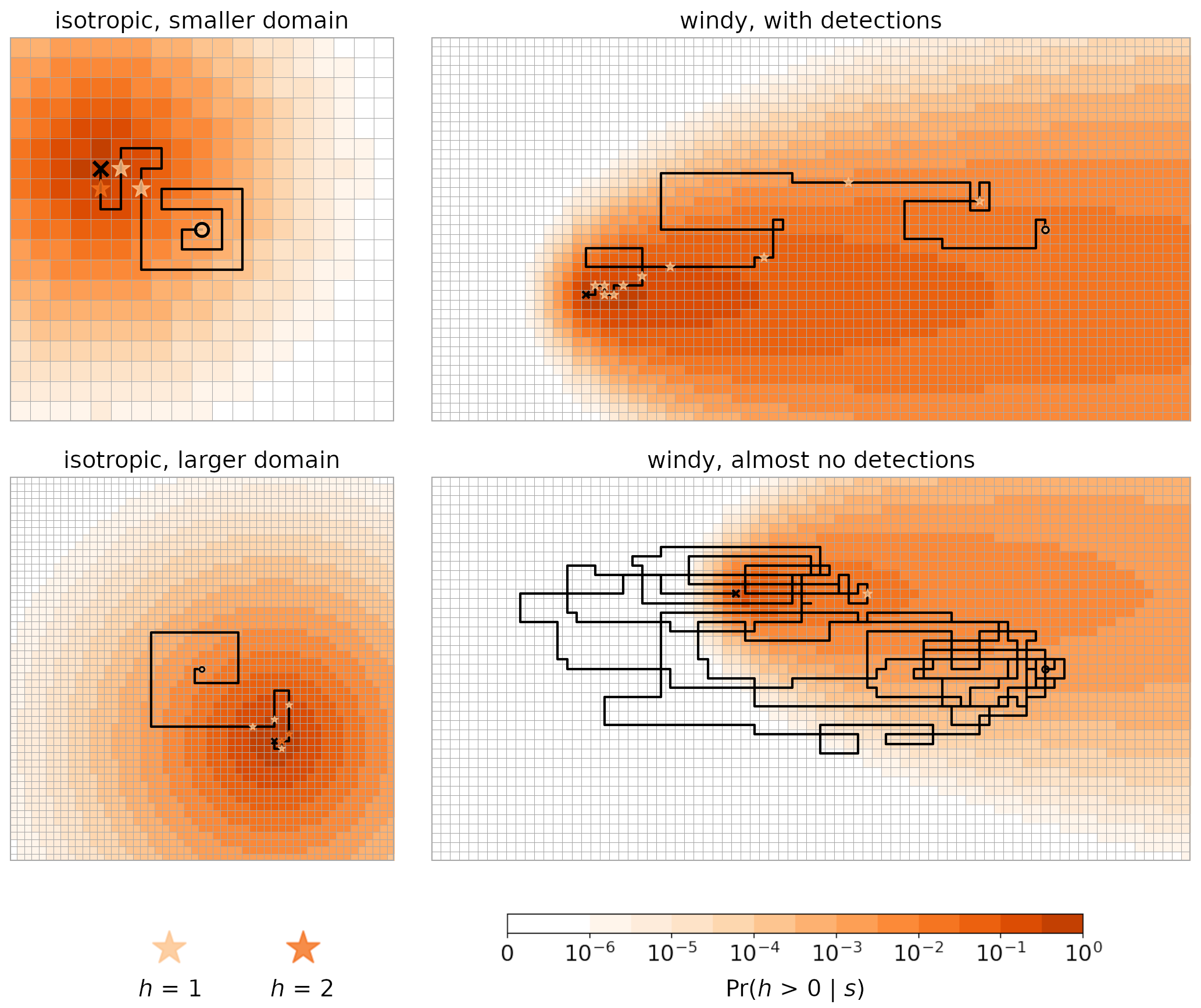}
\caption{Examples of quasi-optimal search trajectories for the four test cases. The agent's starting point and the source's position are denoted by a circle and a cross, respectively. Nonzero hits (odor detections) are indicated by stars. The orange shading color-codes the likelihood of making a detection in each grid cell, given its position with respect to the source. The trajectories were obtained with the best policies for each case.}
\label{fig:trajectory_cases}
\end{figure*}

Overall, all solvers are able to obtain decent policies for all test cases. Examples of quasi-optimal search trajectories are provided in Fig.~\ref{fig:trajectory_cases}. In all cases typical searches roughly consist of an exploratory phase followed by an exploitative phase after the first detections are made. At the beginning of the search, the source can be anywhere within a vast area, and the agent executes long straight moves to explore it efficiently (forming a spiral in the isotropic cases, or long upwind surges in the windy cases). When an odor is detected (nonzero hit), the belief suddenly narrows down to a much smaller area which is very likely to contain the source. The agent then restricts its moves to this area which is searched in a more exhaustive manner.

In the isotropic cases, the main difficulty is the lack of directionality. As the source is initially equally likely to be in any direction, and hits only inform about the distance to the source, the belief typically exhibits a high degree of symmetry around the agent. Committing to a given direction will necessarily incur a large penalty if the source is actually located in any other direction. In the windy cases, the main difficulties are the long distance that typically separates the source and the agent, resulting in longer search times compared to the isotropic cases, and the risk of ``missing'' the source (passing by it upwind) and leaving the odor plume (this explains why the agent goes back twice in Fig.~\ref{fig:trajectory_cases}, top right panel). The windy case with almost no detection is an extreme example where the search is very long but almost deterministic: unlike other cases, here the agent is almost sure to detect nothing at each step.

\begin{figure*}%
\centering
\includegraphics[width=0.99\textwidth]{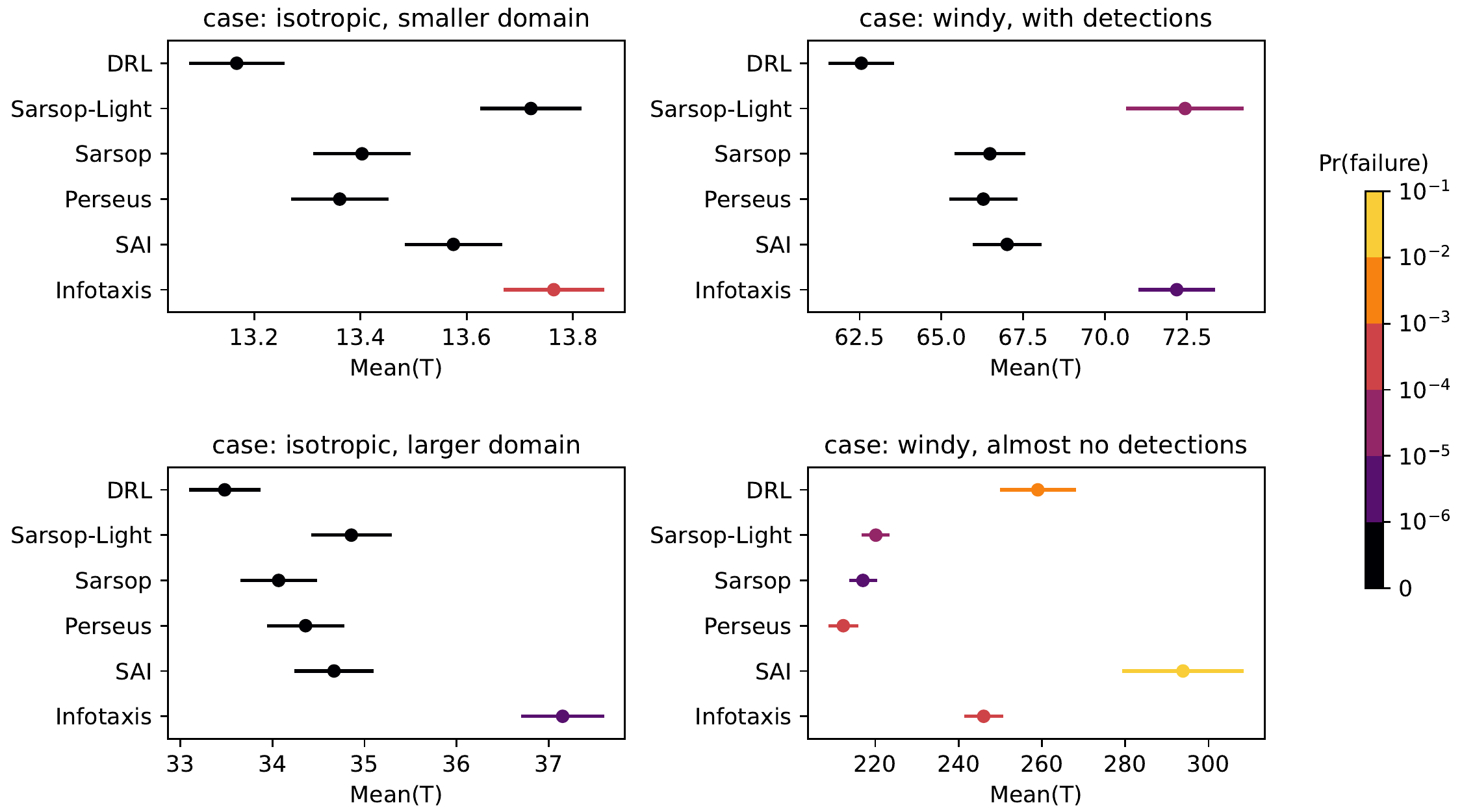}
\caption{Performance of various policies on the four test cases, measured by the mean search time to find the source (lower is better). The probability of never finding the source is color-coded. Four policies have been computed using numerical solvers: deep reinforcement learning (DRL)\cite{Loisy2022a}, Sarsop \cite{Kurniawati2008} and its light version (Sarsop-Light) where the number of $\alpha$-vectors is constrained such that the number of free parameters is comparable to DRL, and Perseus \cite{Spaan2005}. Two heuristic policies are shown for comparison: infotaxis \cite{Vergassola2007} and space-aware infotaxis (SAI) \cite{Loisy2022a}.}
\label{fig:results}
\end{figure*}

The quantitative performance of the various policies on the four test cases is reported in Fig.~\ref{fig:results}. Two metrics are considered: $\Pr(\text{failure})$, the probability of never finding the source, and $\Mean(\text{T})$, the mean time to find the source provided it is ultimately found. Three policies have been computed using numerical solvers: deep reinforcement learning (DRL), Sarsop and Perseus. For comparison, two state-of-the-art heuristic policies are also shown: infotaxis \cite{Vergassola2007} and space-aware infotaxis \cite{Loisy2022a}.

DRL beats Perseus and Sarsop on 3 out of 4 test cases (the two isotropic cases, and the windy case with detections). The probability of never finding the source is negligible for the three solvers, and the mean time to find the source is lower with DRL. Perseus and Sarsop perform comparably well. These results also show that space-aware infotaxis is very close to the optimal performance on these three test cases.

DRL, however, fails at obtaining a quasi-optimal policy for the windy case with almost no detections, where it performs worse than other solvers and than infotaxis (space-aware infotaxis is not a good policy in this case). This test case is particular, as the search is very long but essentially deterministic as odor detections are extremely rare. The failure of DRL in this scenario could be due to inappropriate hyperparameters, though we performed limited testing of those without success. We speculate that a possible explanation is that epsilon-greedy exploration, which we used in our implementation of DRL, is known to be deficient for problems with long time horizons and should be replaced with a form of ``deep exploration'' \cite{Osband2016}. However, this is beyond the scope of the present work.

\subsection{Discussion}
\label{sec:bench_discussion}

While Perseus was found to perform well overall, it is worth reminding the reader that it requires a good heuristic to collect beliefs as well as a clever reward shaping. Here we used infotaxis as a heuristic, which is already close to optimality, and the reward shaping was based on trial and error. Therefore Perseus is actually inferior to DRL and Sarsop in general, which require no domain knowledge or human intuition. In our view, the sole advantage that Perseus enjoys over Sarsop is its applicability to problems with a broad distribution of initial beliefs. In this section we discuss further the pros and cons of Sarsop and DRL.

Beyond the raw performance of the policies on the task, it is interesting to consider the computational cost that each method entails. We will not provide quantitative metrics, as (i) the solver implementations are done in different languages (Python for DRL, C++ for Sarsop) and (ii) our tests have been performed on vastly different machines. But we will provide qualitative remarks on our experience with using these various approaches.

As a solver, DRL is painfully slow compared to Sarsop: DRL can take several days when Sarsop provides a solution within a couple hours. On the other hand, Sarsop provides very large policies which are exceedingly slow to execute. In comparison, the neural network policies obtained by DRL are much lighter and faster to execute, typically by a factor of roughly 10. To make Sarsop competitive with DRL with respect to execution time, we computed additional Sarsop policies, called ``Sarsop-Light'' in Fig.~\ref{fig:results}. Sarsop-Light policies are defined by a smaller number of $\alpha$-vectors, chosen such that the number of free parameters that parameterize the policy is comparable to that of the corresponding DRL policies (cf. Table~\ref{tab:architecture} and Table~\ref{tab:nb_alphavectors}). We found that Sarsop-Light policies are typically inferior to DRL policies.

\section{Conclusion}
\label{sec:conclusion}

In this paper we have compared two computational methods for approximately solving large POMDPs: a recently proposed one based on deep reinforcement learning (DRL), and the standard one relying on point-based value iteration (PBVI). We benchmarked these methods on variants of the olfactory search POMDP, a Goal-POMDP where the agent must find a hidden odor source as quickly as possible using stochastic partial observations in the form of odor cues. As PBVI comes in a number of slightly different flavors, we chose for our benchmark two popular PBVI implementations: Sarsop and Perseus.

While all solvers performed well overall, DRL outperformed PBVI on 3 out of 4 test cases. DRL shines by its ability to provide lighter policies with significantly faster execution speed compared to the policies generated by PBVI. Constraining PBVI solvers to reduce the size of their policies considerably degrade their performance on the task.

On the other hand, obtaining good policies with DRL requires training a large deep neural network, which is a costly process. In comparison, PBVI can generate approximate solutions much more efficiently. DRL also involves a large number of hyperparameters. Even though hyperparameter optimization is not needed to obtain good solutions, some minimal tuning is needed, which further increases the cost of training. Finally, on the 4th test case (a very long, almost deterministic search), the performance of DRL is significantly worse than that of PBVI for reasons that remain to be elucidated.

To summarize, DRL is competitive with respect to traditional PBVI solvers, and the best choice depends on the use case. PBVI solvers are best if the computation time allowed to the solver is limited (and we recommend Sarsop over Perseus). DRL is best if the execution time or the policy size is limited, as is usually the case in robotics. Solutions to the olfactory search POMDP have applications to sniffer robots, and DRL offers interesting perspectives for the future of these robots.

DRL also has a strong advantage in its flexibility: it can be applied without convergence issue to undiscounted problems, whereas point-based solvers require a strictly less-than-unity discount factor. For this reason, DRL has potential to be applied to a risk-sensitive setting by appropriately generalizing the Bellman equation \cite{Marcus1997,Coraluppi1999}, an idea which will be explored in future research.

The olfactory search POMDP we considered is model-based: the agent knows the process that generates odor detections (the ``model''), which allows it to maintain a belief using Bayesian inference. This assumption  can be relaxed by considering instead an agent that makes decisions based directly on its current observation and some internal memory state, typically using a recurrent neural network \cite{Singh2023}. 
In principle, the model-free agent can attain the same performance on the task as the model-based one. In practice, recurrent neural networks are hard to train and it will be interesting to evaluate their performance in the light of the (near-)optimal performance that can be computed in the model-based setting.

\backmatter

\bmhead{Acknowledgments}
We thank Luca Biferale, Antonio Celani, Massimo Vergassola, and Christophe Eloy for useful discussions.
AL received funding from the European Research Council (ERC) under the European Union's Horizon 2020 research and innovation programme (grant agreement No 834238). 
RAH received funding from the European Union’s H2020 Program under grant agreement No. 882340.
Centre de Calcul Intensif d'Aix-Marseille is acknowledged for granting access to its high performance computing resources.

\section*{Statements and Declarations}

\subsection*{Competing interests}
The authors have no competing interests to declare.

\subsection*{Code and data availability}
The code used to generate DRL policies and to evaluate all policies on the olfactory search POMDP is available at \url{https://github.com/auroreloisy/otto-benchmark}. The policies computed with the various solvers can be downloaded at \url{https://doi.org/10.5281/zenodo.7586357}. The data used to plot the results in Fig.~6 is available at \url{https://doi.org/10.5281/zenodo.7586312}.
PBVI policies were obtained using the code available at \url{https://github.com/rheinonen/PerseusPOMDP/} for Perseus, and at \url{https://github.com/rheinonen/sarsop/} for Sarsop.

\subsection*{Authors' contributions}
AL and RAH designed the study, performed the research and analyzed the results. AL wrote the manuscript with the help of RAH.

\begin{appendices}

\section{Observation model: odor dispersion and detection in turbulence}
\label{app:obs_model}

In this appendix we provide the model used to generate observations (hits), that is, we specify $\Pr(h \mmid s)$. This model is based on a physical modeling of odor dispersion and detection in a turbulent flow.

We consider a point source that emits, at a rate $R$, odor particles with a finite lifetime $\tau$. They disperse in the turbulent medium, which is characterized by an effective diffusivity $D$ and a mean wind speed $V$. The wind blows in the positive x-direction, denoted by the unit vector $\v{e}_x$. The searcher is modeled as a sphere (disk) of diameter $\Delta x$ fully covered in receptors. A every step, it takes a ``sniff'' over a time $\Delta t$, during which odor particles diffusing to its surface are absorbed. It then moves by one body length $\Delta x$ (which defines the size of a grid cell).

Based on these assumptions, one can derive the mean number $\mu$ of odor particles (``hits'') detected by the searcher as a function of its position $\v{r}$ with respect to the source \cite{Vergassola2007,Loisy2022a}:
\begin{subequations}
\begin{align}
 \mu(\v{r}) & = \frac{R \Delta t}{\ln(2 \lambda/\Delta x)} \exp \left( \frac{V \v{r} \cdot \v{e}_x}{2D}\right) K_{0} \left( \frac{\abs{\v{r}}}{\lambda} \right) & \text{in 2D} \label{eq:mu_2D} \\[0.5em]
 \mu(\v{r}) & = \frac{R \Delta t \Delta x }{2 \abs{\v{r}}} \exp \left( \frac{V \v{r} \cdot \v{e}_x}{2D} - \frac{\abs{\v{r}}}{\lambda} \right) & \text{in 3D} \label{eq:mu_3D}
\end{align}
\end{subequations}
with
\begin{equation}
 \lambda = \sqrt{ \frac{D \tau}{1+ \dfrac{V^2 \tau}{4 D}}}
\end{equation}
and where $K_0$ is the modified Bessel function of the second kind of order 0.
The number of hits is distributed according to a Poisson's law
\begin{equation}
 \Pr(h \mmid \mu) = \frac{\mu^h \exp(-\mu)}{h!}
\end{equation}
with mean $\mu$.

This completes the description of the observation model $\Pr(h \mmid s)$. In the main text, $\v{r}$ is denoted $s$ to be consistent with standard POMDP notations. The likelihood of observing $h$ in state $s$ corresponding to position $\v{r}$ is given by
\begin{equation}
 \Pr(h \mmid s) = \Pr(h \mmid \mu(\v{r})).
\end{equation}

\section{Further specifications of the POMDP variants used as test cases}
\label{app:cases_details}

\subsection{Search initialization}
\label{app:init}

The search start by drawing a initial belief $b_0$ from the set of initial beliefs $\mathcal{B}_0$. Here we provide further details about the construction of this set.

Before the search starts, we assume that the searcher is located in an infinite domain containing a source which location is distributed uniformly. We calculate the probabilities of detecting an odor, $\Pr(h_0)$ for $h_0>0$, based on the observation model: $\Pr(h) = \sum_s \Pr(h \mmid s) \Pr(s)$ where $\Pr(s)$ is a constant here. Then, we initialize the finite-size grid with a uniform prior $b_{-1}(s)$. We draw a nonzero hit from the distribution $\Pr(h_0)$, and perform the Bayes update of the belief accordingly. This gives us $b_0$. The search starts from this $b_0$. Each initial nonzero hit value $h_0$ yields a different initial belief $b_0$. Effectively, the initial belief is drawn randomly from the set of possible initial beliefs $\mathcal{B}_0$ built from the possible values of $h_0$. The reader is referred to \cite{Loisy2022a} for more details.

The set of initial beliefs and their probabilities depend on the test case, cf. Table~\ref{tab:initial_hit}. In the isotropic cases, it contains several elements. In the windy cases, hits are binary ($h=1$ for a detection, $h=0$ otherwise) so a single initial belief is considered.

\begin{table}
\begin{center}
\begin{tabular}{@{}lllll@{}}
\toprule
case & $\Pr(h_0=1)$ & $\Pr(h_0=2)$ & $\Pr(h_0=3)$ & $\abs{\mathcal{B}_0}$\\
\midrule
isotropic, smaller domain    &  $0.85$ & $0.15$ & - & 2 \\
isotropic, larger domain    & $0.83$ & $0.13$ & $0.04$ & 3 \\
windy, with detections & $1.00$ & - & - & 1 \\
windy, almost no detections & $1.00$ & - & - & 1\\
\botrule
\end{tabular}
\caption{Search initialization: observation values (number $h$ of hits) and their probabilities of occurring. Each value of $h_0$ is used to generate an initial belief $b_0$. The set $\mathcal{B}_0$ of initial beliefs contains as many elements as there are possible observation values. The probability that a particular $b_0$ is drawn at the beginning of a search is given by the associated $\Pr(h_0)$.}
\label{tab:initial_hit}%
\end{center}
\end{table}

\subsection{Observation model}

In the two isotropic cases, $V=0$ and we use the 2D version of the observation model (Eq.~\ref{eq:mu_2D}), which reduces to:
\begin{equation}
 \mu(\v{r}) = \frac{R \Delta t}{\ln(2 \lambda/\Delta x)} K_{0} \left( \frac{\abs{\v{r}}}{\lambda} \right)
\end{equation}
with $\lambda = \sqrt{D \tau}$. It is fully defined by specifying two dimensionless parameters, which we chose as $\lambda/\Delta x$ and $R \Delta t$, as was done in \cite{Loisy2022a}. Their numerical values are given in Table~\ref{tab:isotropic_values}.

In the two windy cases, we use the 3D version of the observation model (Eq.~\ref{eq:mu_3D}). To fully specify the model, three dimensionless parameters must be set. For consistency with \cite{Heinonen2022arxiv} we choose
\begin{equation}
 \bar{R} = \frac{R \Delta t}{2} \qquad
 \bar{V} = \frac{V \Delta x}{D} \qquad
 \bar{\tau} = \frac{V^2 \tau}{D}.
\end{equation}
Their values are provided in Table~\ref{tab:windy_values}.

\begin{table}
\begin{center}
\begin{tabular}{@{}lccc @{}}
\toprule
case & $\lambda/\Delta x$ & $R \Delta t$ \\
\midrule
isotropic, smaller domain & 1.0 & 1.0 \\
isotropic, larger domain & 3.0 & 2.0 \\
\botrule
\end{tabular}
\caption{Parameters of the observation model used in the isotropic cases (no wind).}
\label{tab:isotropic_values}%
\end{center}
\end{table}

\begin{table}
\begin{center}
\begin{tabular}{@{}lcccc@{}}
\toprule
case & $\bar{R}$ & $\bar{V}$ & $\bar{\tau}$ \\
\midrule
windy, with detections & 2.5 & 2 & 150 \\
windy, almost no detections & 0.25 & 2 & 150 \\
\botrule
\end{tabular}
\caption{Parameters of the observation model used in the windy cases.}
\label{tab:windy_values}%
\end{center}
\end{table}

\section{Methodological details on DRL, Sarsop and Perseus}
\label{app:solvers}

\subsection{Deep reinforcement learning}
\label{app:solver_drl}

\begin{algorithm}
\caption{Reinforcement learning algorithm used to train a neural network to approximate the optimal value function.}\label{alg:drl}
\algrenewcommand{\algorithmiccomment}[1]{\#\ #1}
{\small
\begin{algorithmic}[0]
\State Initialize replay memory to capacity \textsf{memory\_size}
\State Initialize value function $v$ with random weights $w$
\State Initialize target value function $v^-$ with random weights $w^- = w$
\State $it \leftarrow 0$
\Repeat
    \State \Comment{Generate new experience}
    \State $epsilon \leftarrow \max(\textsf{epsilon\_init} * \exp(-it/\textsf{epsilon\_decay}), \textsf{epsilon\_floor})$ \Comment{decaying $\epsilon$}
    \State $m \leftarrow 0$
    \State $episode\_complete \leftarrow$ True
    \While{$m < \textsf{new\_transitions\_per\_it}$}
        \If{episode\_complete}
            \State initialize belief $b$ for a new episode
            \State $episode\_complete \leftarrow$ False
        \EndIf
        \State $b \leftarrow$ apply\_random\_symmetry($b$) \Comment{randomize over symmetries}
        \State for all actions $a$, compute all $b'$ accessible from $b$
        \State store $(b,a,b')$ in replay memory
        \State $m \leftarrow m+1$
        \State with probability $epsilon$ select a random action $a$, \Comment{$\epsilon$-greedy exploration}
        \State otherwise select action $a =\argmax_a \sum_{b'} \Pr(b'\mmid b,a) v(b';w)$ 
        \State $b \leftarrow$ make\_step\_in\_env($b$, $a$) \Comment{execute action and transition to a new belief}
        \State $episode\_complete$ $\leftarrow$ $b = b^\Omega$
    \EndWhile
   \State \Comment{Update weights by stochastic gradient descent}
    \For{$gd\_step=1,\textsf{gd\_steps\_per\_it}$}
    \State Sample $\textsf{minibatch\_size}$ transitions $(b,a,b')$ from replay memory
    \State For each transition, compute targets $y$ using the delayed target network:
    \State $y = -1 + \max_a \sum_{b'} \Pr(b' \mmid b,a) v^-(b';w^-)$
    \State Perform a gradient descent step on $\left(y - v(b;w) \right)^2$ w.r.t network weights $w$
    \EndFor
    \State $it \leftarrow it + 1$
    \State every $\textsf{update\_target\_network\_it}$ iterations, reset $v^- = v$
\Until{weights $w$ have converged}
\end{algorithmic}
}
\end{algorithm}

To approximate the optimal value function, we use a fully connected neural network with 3 hidden layers. Extensive experiments by \cite{Loisy2022a} suggest that as a rule of thumb, one should choose a number of neurons per layer roughly proportional to the size $\abs{\mathcal{S}}$ of the input (the belief). Here we used either 512 or 1024 neurons per layer depending on the test case, cf. Table~\ref{tab:architecture}.

For a search domain of size $n_x \times n_y$, the belief is a two-dimensional array of probability values with size $(2 n_x-1) \times (2 n_y - 1)$.
Each entry corresponds to a possible position of the source with respect to the agent: if we denote the coordinates of the center of the array as ($0$, $0$), the source's possible position relative to the agent ranges from $-(n_x-1)$ to $(n_x -1)$ along the x-direction, and from $-(n_y-1)$ to $(n_y -1)$ along the y-direction. 
Since the belief size is four times larger than the grid size, 3/4th of the entries are outside the search domain and are zeros (cf. Fig.~\ref{fig:nn_illustration}). This two-dimensional array is flattened before being fed to the network.

The pseudo-code of the reinforcement learning algorithm is provided in Algorithm~\ref{alg:drl}. The hyperparameters we used are given in Table~\ref{tab:hyperparameters}. They were not optimized (it would be too costly) but chosen based on \cite{Loisy2022a} and limited experiments which showed that the learned policy is essentially insensitive to hyperparameters (provided they are given reasonable values).

\begingroup
\setlength{\tabcolsep}{5pt}
\begin{table}
\begin{center}
\begin{tabular}{@{}lrrr@{}}
\toprule
case & input size  & hidden layers & number of free parameters \\
\midrule
isotropic, smaller domain    & 1369 & 3$\times$512 & 1,227,265 \\
isotropic, larger domain    & 11025 & 3$\times$1024 & 13,390,849 \\
windy, with detections & 13041 & 3$\times$1024 & 15,455,233 \\
windy, almost no detections & 13041 & 3$\times$1024 & 15,455,233 \\
\botrule
\end{tabular}
\caption{Size of the fully connected neural network used for each case. We always use 3 hidden layers. The number of free parameters (weights) is $N=H(I+2H+4)+1$ where $I$ is the input size (number of states $\abs{\mathcal{S}}$) and $H$ the number of neurons per hidden layer.}
\label{tab:architecture}%
\end{center}
\end{table}
\endgroup

\begin{table}
\begin{center}
\begin{tabular}{lll}
\toprule
hyperparameter & value & description \\
\midrule
learning rate & 0.001 & for stochastic gradient descent (SGD) \\
\textsf{epsilon\_init} & 1.0 & initial value of $\epsilon$ for $\epsilon$-greedy exploration \\
\textsf{epsilon\_floor} & 0.1 & final value of $\epsilon$ for $\epsilon$-greedy exploration \\
\textsf{epsilon\_decay} & 20000 & time scale for decay of $\epsilon$ \\
\textsf{memory\_size} & 1000 & number of transitions stored in memory \\
\textsf{minibatch\_size} & 64 & size of the mini-batch for SGD updates \\
\textsf{new\_transitions\_per\_it} & 192 & transitions added to memory per iteration \\
\textsf{gd\_steps\_per\_it} & 12 & number of SGD updates per iteration \\
\textsf{update\_target\_network\_it} & 1 & frequency of target network updates \\
\botrule
\end{tabular}
 \caption{List of hyperparameters (as usually defined or defined in Algorithm~\ref{alg:drl}) and the typical values we used. Actual values used for each case may slightly differ, refer to our implementation at \url{https://github.com/auroreloisy/otto-benchmark}.}
 \label{tab:hyperparameters}
\end{center}
\end{table}

\subsection{PBVI algorithms}
\label{app:solver_pbvi}

The olfactory search POMDP is an example of POMDP with undiscounted rewards. More generally, one can introduce a discount factor $\gamma \in (0,1]$ and the cumulated reward for an episode is $\sum_{t=1}^{T} \gamma^{t-1} r_t$. The original problem corresponds to $\gamma = 1$. Using $r_t=-1$, we have:
\begin{equation}
 \mathbb{E}\left[\sum_{t=1}^T \gamma^t r_t \right] =
 \begin{cases}
 -\mathbb{E}[T] & \quad \text{for $\gamma = 1$}, \\[1em]
 -\dfrac{1-\mathbb{E}[\gamma^T]}{1-\gamma} & \quad \text{for $\gamma < 1$}.
\end{cases}
\end{equation}

The discount factor helps regularize the problem by suppressing the influence of times far in the future on the policy. Formally, each choice of $\gamma$ defines a different POMDP, but presently we will continue to treat the (undiscounted) mean arrival time as our objective function and consider $\gamma$ to be a tunable hyperparameter, necessary for the functioning of most popular POMDP algorithms (Sarsop and Perseus included). Generally, we try to keep it as close to unity as possible; for Sarsop, we chose $\gamma=0.98,$ and for Perseus we chose one of 0.95, 0.96 or 0.98, tuned by hand to optimize performance.

\begingroup
\setlength{\tabcolsep}{2pt}
\begin{table}
\begin{center}
\begin{tabular}{@{}lrrrrrr@{}}
\toprule
& \multicolumn{6}{c}{number of $\alpha$-vectors (number of free parameters)} \\[0.5em]
case & \multicolumn{2}{c}{Sarsop}  & \multicolumn{2}{c}{Sarsop-Light} & \multicolumn{2}{c}{Perseus} \\
\midrule
isotropic, smaller domain    	& 19,834 & (27,152,746) & 880 & (1,204,720) & 2,058 & (2,817,402) \\
isotropic, larger domain    	& 10,523 & (116,016,075) & 1,207 & (13,307,175) & 4,392 & (48,421,800) \\
windy, with detections 			& 9,314 & (121,463,874) & 1,149 & (14,984,109) & 3,880 & (50,599,080) \\
windy, almost no detections 	& 9,509 & (124,006,869) & 1,149 & (14,984,109) & 670 & (8,737,470) \\
\botrule
\end{tabular}
\caption{Size of the PBVI policy, for each case. The number of free parameters is equal to the number of $\alpha$-vectors multiplied by the number of states $\abs{\mathcal{S}}$.}
\label{tab:nb_alphavectors}%
\end{center}
\end{table}
\endgroup

In order to obtain competitive policies with Perseus, we usually found it necessary to transform the problem by introducing a potential shaping function \cite{Ng1999} to the reward. Defining $r(b,a)$ as the reward obtained when executing action $a$ in belief $b$, one can show that replacing the original (constant) reward function $r(b,a) = -1$ by the shaped reward function
\begin{equation}
 r(b,a) = -1 + F(b,a)
\end{equation}
with $F(b,a)$ a function of the form
\begin{equation}
F(b,a) = \phi(b) - \gamma \sum_{b'} \Pr(b' \mmid b,a) \phi(b'),
\end{equation}
preserves the optimal policy.
A clever choice can sometimes accelerate convergence of value iteration. We took
\begin{equation}
\phi(b) = - c \sum_{s \in \mathcal{S}} D(s) \cdot b(s)
\end{equation}
with $c$ constant and $D(s)$ the Manhattan distance between the agent and the source in state $s$. The point of this choice is to incentivize the agent to move closer to the source. The hyperparameter $c$ was tuned from problem to problem.

We found that policies do not always improve monotonically under Perseus, so policies were evaluated empirically after every iteration, and the algorithm was terminated when the mean arrival time failed to improve for some selected number of iterations.

As written, the Sarsop algorithm takes as input a single prior which serves as the root of the belief tree; Perseus, in contrast, accepts an arbitrary distribution of priors. For the isotropic problems, where several priors are possible, we simply used the most likely prior, corresponding to a single detection ($h_0=1$) at time $t=0$ (cf. Appendix~\ref{app:init}).

The size of the PBVI policies is reported in Table~\ref{tab:nb_alphavectors}.

\end{appendices}



\end{document}